# Context-Aware Monolingual Human Evaluation of Machine Translation

**Silvio Picinini**
eBay
spicinini@ebay.com

**Sheila Castilho**
Salis/Adapt Centre
Dublin City University
sheila.castilho@dcu.ie

## Abstract

This paper explores the potential of context-aware monolingual human evaluation for assessing machine translation (MT) when no source is given for reference. To this end, we compare monolingual with bilingual evaluations (with source text), under two scenarios: the evaluation of a single MT system, and the comparative evaluation of pairwise MT systems. Four professional translators performed both monolingual and bilingual evaluations by assigning ratings and annotating errors, and providing feedback on their experience. Our findings suggest that context-aware monolingual human evaluation achieves comparable outcomes to human bilingual evaluations, and suggest the feasibility and potential of monolingual evaluation as an efficient approach to assessing MT.

## 1 Introduction

MT evaluation has traditionally depended on comparing the translated text with its corresponding source text to assess the MT performance. However, several works in the area have explored the potential of monolingual evaluation (Callison-Burch, 2005; Koehn, 2010; Mitchell et al., 2013; Schwartz, 2014; Fomicheva and Specia, 2016; Graham et al., 2017).

Building on earlier studies suggesting that monolingual post-editors, even without access to the source text, can enhance MT output quality (Callison-Burch, 2005; Koehn, 2010; Mitchell et al., 2013), Schwartz (2014) examined the effectiveness of monolingual post-editing performed by domain experts in improving MT quality and reducing reliance on bilingual post-editing. The author reported that a subject-matter expert monolingual post-editor confidently corrected 87% of sentences based solely on the target text, with 96% of these corrections deemed appropriate following bilingual verification. While our work presented here differs in focus, exploring monolingual evaluation rather than post-editing, it highlights a complementary strength: the importance of context-awareness.

The work of Graham et al. (2017) investigated various evaluation approaches with monolingual evaluation, such as the inclusion or exclusion of reference translations; the impact of additional contextual information (e.g., surrounding sentences); and the influence of displaying metadata to annotators. They concluded that monolingual evaluations, even when crowd-sourced, could effectively measure MT system performance.

Recent advancements in context-aware evaluation methodologies have introduced a deeper focus on the appropriateness and coherence of translations within broader textual contexts (Castilho, 2020, 2021; Castilho et al., 2020; Freitag et al., 2021). These approaches challenge traditional evaluation methodologies that rely on isolated sentences, highlighting the limitations of sentence-level assessments. By integrating contextual information, evaluators can better capture nuanced aspects of translation quality, such as consistency, discourse coherence, and appropriateness within the larger narrative.

Building on this, our study investigates the potential of monolingual assessments in the context, addressing the following research question: *Are monolingual assessments of MT comparable to bilingual assessments?* To explore this, we examine two key scenarios:
a) Can context-aware monolingual assessments of a single MT output provide results comparable to bilingual assessments?
b) Can context-aware monolingual assessments of two different MT outputs (pairwise comparison) achieve comparability with bilingual assessments?

The research question with the two scenarios are illustrated in figure 1. A positive answer to the question would open potential benefits for facilitating the evaluation process.

|            | Single MT   | Pairwise MT | Test Set |
|------------|-------------|-------------|----------|
| DELA       | 23 (303 w)  | 30 (576 w)  | 1, 3     |
| Customer Support | 18 (429 w) | 53 (750 w) | 2, 4   |

Table 1: Number of sentences and word count from each source per task, along with their associated Test Sets.

## 2 Methodology

### 2.1 Test Sets

This evaluation was performed using a sentence-in-context format, where the evaluation is at the sentence level, but the evaluator has access to the full context of the document. In one content (travel review), full content is one entire review by a traveler. In the other content (customer support help pages), full content is one entire page explaining a topic to a customer (for example, "listing policies" or "how to buy as a guest"). For this, we used two documents from the Review section of the DELA corpus (Castilho et al., 2021) and two documents from Anonymized's Customer Support pages.[1] The DELA Corpus is a document-Level corpus, in Brazilian Portuguese (the target language that we use in this work) annotated with context-related issues such as ellipsis, gender, lexical ambiguity, number, reference, and terminology. The corpus covers six domains and we chose Reviews (in our case, travel reviews). We constructed 4 different Test Sets. For the scenario where a single MT system output is assessed (Single MT) we used 41 sentences (732 words), 23 from DELA (Test Set 1) and 18 from eBay's Customer Support pages (Test Set 2). For the scenario where two MT systems are assessed (pairwise MT), we used 83 sentences (1326 words), 30 from DELA (Test Set 3) and 53 from eBay (Test Set 4). Number of sentences and word counts are in table 1.

### 2.2 Systems, Evaluators and Language

The MT systems used for the evaluation were the freely available Microsoft (MT1) and DeepL (MT2). We note that only MT1 (Microsoft) is used in the Single MT scenario.[2]

The language pair of this experiment was English (EN) as the source and Brazilian-Portuguese (pt-BR) as the target. Four in-house professional translators participated in the evaluation. They are native speakers of pt-BR and long-time translators for eBay and for a major language services provider.

### 2.3 Metrics

Evaluators assessed the systems' output in terms of:

- Overall Ratings (Likert scale from 1-5)
- MQM Error Annotation

**Ratings:** these Ratings, used at eBay in-house for MT performance evaluation, is measured in a 1-5 scale and considers both adequacy and fluency. It is used to evaluate the experience and success of an end-user in understanding the final MT output, and also the experience of a translator who receives the MT output as an input for post-editing. The reason for using this in-house Ratings is to maintain ecological validity, as it reflects the Ratings these translators are familiar with. Additionally, since eBay will continue using these Ratings in future evaluations, the results here can provide valuable insights for the company's ongoing comparisons.

The scale consists of:

1. Critical/Incomprehensible: When the translation:
   - is incomprehensible (hallucinations, meanings that make no sense in the text)
   - is completely not in the target language
   - contains critical errors
   - can be a misleading mistranslation
   - may carry health, safety, legal, reputation, religious or financial implications
   - contains profanities/insulting words not appropriate in the context

2. Significant errors/effort: Translation contains significant errors and may be incomplete (such as significant mistranslations, significant portions untranslated, significant omissions) and requires significant effort to understand it. It would require a significant amount of change to be usable.

---

[1] Document 1 is "Buying as a Guest", available at url anonymized, and document 2 is "Listing Policies", available at url anonymized

[2] MT translations were obtained on 22 February 2024 from the public versions of the MT providers on their sites.

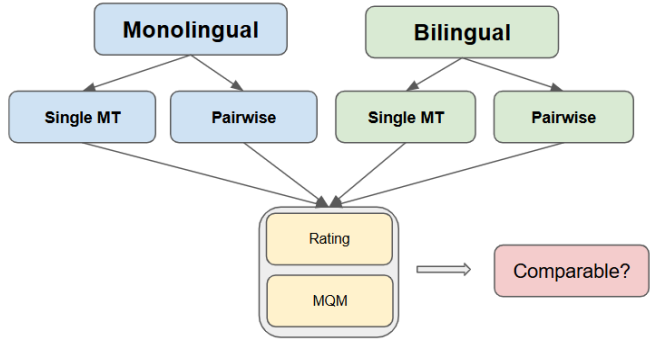

Figure 1: Design of the tasks to answer the RQ. "Single MT" refers to the scenario where only one MT output is assessed, whereas "Pairwise" refers to the scenario where two MT systems are assessed. Both scenarios are assessed in the two modalities, monolingual and bilingual, in terms of Ratings and Error Annotation.

3. Some errors/some effort: There are errors in translation, such as untranslated words and mistranslations (wrong but not misleading translations). The translation requires some effort to understand it and change it.

4. Do not affect understanding: There are minor grammar errors: wrong agreement, word order, missing prepositions, minor omissions or other errors that do not affect understanding. It would require only minor changes.

5. Complete/fluent: Translation is (or seems to be) correct, complete, fluent, easy to read, and contains no errors. It looks like (almost) human translation and requires no changes.

**Error Annotation:** Evaluators annotated errors using the MQM error typology (Lommel et al., 2014), following the methodology and guidelines outlined in (Freitag et al., 2021) (see Appendices A and B).

### 2.4 Setup

The evaluation was conducted using an online spreadsheet sent to the evaluators. As mentioned earlier, the assessment was performed in context, with access to the full text in two modalities:
a) ***Monolingual***, where only the MT output was provided - Task 1.
b) ***Bilingual***, where both the MT output and source text were provided - Task 2.

Evaluators performed both Ratings evaluations and Error Annotations (see Section 2.3) across these two modalities (Monolingual and Bilingual), in two scenarios: Single MT and Pairwise MT. This resulted in a total of 16 evaluations.

The evaluation process followed a structured sequence, with Task 1 (Monolingual) always performed first, followed by Task 2 (Bilingual), with a one-week interval between tasks. This order was designed to minimize the likelihood of evaluators recalling source text information from the bilingual evaluation. By conducting the monolingual evaluation first, evaluators identified visible errors in the target text without reference to the source. In the subsequent bilingual evaluation of the same text, evaluators could incorporate errors previously identified during the monolingual assessment and, additionally, detect new errors that were only apparent through comparison with the source text. The goal of this approach was to determine how many errors identified during the bilingual evaluation had already been noted in the monolingual assessment and how many new errors were detected only through bilingual evaluation.

Moreover, we note that the order of Test Sets and scenarios were randomized. The distribution and randomization of the evaluations can be seen in table 2.

## 3 Results

This section presents the outcomes of the experiments, organized by the tasks performed. The first part, Section 3.1, details the results from Task 1, carried out in the **Monolingual** modality, while Section 3.2 focuses on Task 2, conducted in the **Bilingual** modality. The results for both tasks are reported across the two scenarios - *Single MT* and *Pairwise MT* - covering the overall Ratings and Error Annotation.

| Modality | Evaluator 1 | | Evaluator 2 | | Evaluator 3 | | Evaluator 4 | |
|---|---|---|---|---|---|---|---|---|
| | Scenario | Test Set | Scenario | Test Set | Scenario | Test Set | Scenario | Test Set |
| **Task 1 - Monolingual** | Pairwise | TS3 | Pairwise | TS4 | Single | TS1 | Single | TS2 |
| | | TS4 | | TS3 | | TS2 | | TS1 |
| | Single | TS1 | Single | TS2 | Pairwise | TS3 | Pairwise | TS4 |
| | | TS2 | | TS1 | | TS4 | | TS3 |
| **Task 2 - Bilingual** | Single | TS2 | Single | TS1 | Pairwise | TS4 | Pairwise | TS3 |
| | | TS1 | | TS2 | | TS3 | | TS4 |
| | Pairwise | TS4 | Pairwise | TS3 | Single | TS2 | Single | TS1 |
| | | TS3 | | TS4 | | TS1 | | TS2 |

Table 2: Distribution of evaluation tasks by modality (Monolingual - Task 1, and Bilingual - Task 2), scenarios (Single MT and Pairwise MTs), and Test Sets (TS) assigned to each evaluator.

### 3.1 Task 1 - Monolingual Evaluation

The monolingual results for the Single MT scenario are shown in table 3, while the results for the Pairwise MT scenario are shown in table 4.

**Overall Ratings -** The overall Ratings for the *Single MT scenario* show a significant variation among evaluators (table 3). Interestingly, results for the *Pairwise MT scenario* show a more consistent Ratings for the MT2. For MT1, there is a wider range of Ratings, with evaluator 4 penalizing MT1 more severely (table 4).

**Error Annotation -** The results for the Error Annotation are reported according to i) the overall number of errors, and ii) the severity Major. We provide the number of Minor errors only to facilitate the understanding that All Errors is the sum of Major and Minor errors. Looking at i) the overall number of errors, we are assessing the overall ability of the monolingual evaluation of capturing as many of the total errors as the bilingual evaluation. Looking only at Major errors, we incorporate the severity element into the analysis and assess the ability of monolingual evaluation capturing the important errors, in the same way that a bilingual evaluation would.

**i) All errors -** *Single MT scenario*: We note a wide variation in the reported numbers from evaluators 3 and 4, with evaluators 1 and 2 more in agreement (table 3).
*Pairwise MT scenario*: Also shows a wide variation in the number of total errors for MT2 and for MT1 (table 4).

**ii) Major errors -** *Single MT scenario*: When only Major errors are considered, a wide variation can be seen from the same evaluators in the Single MT scenario.

*Pairwise MT scenario*: Similarly, when only Major errors are considered when evaluating two MTs monolingually, we see for MT2 a fair variation, whereas for MT1 a wider variation is observed.

### 3.2 Task 2 - Bilingual Evaluation

This subsection presents the results for the Bilingual modality, covering both the overall Ratings evaluation and Error Annotation. The bilingual results for the Single MT scenario are shown in table 5, while results for the Pairwise scenario are shown in table 6.

**Overall Ratings -** We note that Ratings for the *Single MT scenario* in the bilingual modality show a fair variation among evaluators (table 5). *The Pairwise MT scenario* results show more consistent Ratings for the MT2. For MT1, there is a wider range of Ratings (table 6).

**Error Annotation -** **i) All errors -** *Single MT scenario*: We note a wide variation in the reported numbers from evaluators 3 and 4, with evaluators 1 and 2 more in agreement (table 5).
*Pairwise MT scenario*: Also shows a wide variation in the number of total errors for MT2 and for MT1 (table 6).

**ii) Major errors -** *Single MT scenario*: When only Major errors are considered, a wide variation can be seen from the same evaluators in the Single MT scenario.
*Pairwise MT scenario*: Similarly, when only Major errors are considered when evaluating two MTs bilingually, we see for MT2 a fair variation, whereas for MT1 a wider variation is observed.

## 4 Analysis of the Results

The aim of this evaluation is to compare the extent to which MT systems can be assessed monolin-gually to the same (or some) extent as bilingually. This section analyses whether the results of these experiments answer our RQ in the two scenarios explored: Single MT and Pairwise MT.

| Task 1 Single MT | Ratings | Error Annotation | | |
|---|---|---|---|---|
| | | All errors | Major | Minor |
| Eval 1 | 3.8 | 23 | 8 | 15 |
| Eval 2 | 3.7 | 20 | 9 | 11 |
| Eval 3 | 4.4 | 6 | 2 | 4 |
| Eval 4 | 2.9 | 46 | 26 | 20 |
| Average | 3.7 | 24 | 11 | 13 |

Table 3: Task 1 - Monolingual results for Overall Ratings and Error Annotation for the **Single MT** scenario.

| Task 1 Pairwise MT | Ratings | | Error Anotation | | | | | |
|---|---|---|---|---|---|---|---|---|
| | MT1 | MT2 | MT1 | | | MT2 | | |
| | | | All errors | Major | Minor | All errors | Major | Minor |
| Eval 1 | 4.0 | 4.9 | 44 | 28 | 16 | 9 | 3 | 6 |
| Eval 2 | 4.2 | 4.8 | 29 | 13 | 16 | 10 | 2 | 8 |
| Eval 3 | 4.1 | 4.7 | 28 | 20 | 8 | 10 | 4 | 6 |
| Eval 4 | 3.0 | 4.4 | 71 | 45 | 26 | 30 | 10 | 20 |
| Average | 3.8 | 4.7 | 43 | 27 | 17 | 15 | 5 | 10 |

Table 4: Task 1 - Monolingual results for Overall Ratings and Error Annotation for the **Pairwise MT** scenario.

| Task 2 Single MT | Rating | Error Annotation | | |
|---|---|---|---|---|
| | | All errors | Major | Minor |
| Eval 1 | 4.2 | 23 | 6 | 17 |
| Eval 2 | 3.6 | 26 | 7 | 19 |
| Eval 3 | 4.0 | 17 | 9 | 8 |
| Eval 4 | 3.3 | 48 | 15 | 33 |
| Average | 3.8 | 29 | 9 | 19 |

Table 5: Task 2 - Bilingual results for Overall Ratings and Error Annotation for the **Single MT** scenario.

| Task 2 Pairwise MT | Ratings | | Error Annotation | | | | | |
|---|---|---|---|---|---|---|---|---|
| | MT1 | MT2 | MT1 | | | MT2 | | |
| | | | All errors | Major | Minor | All errors | Major | Minor |
| Eval 1 | 4.1 | 4.9 | 41 | 27 | 14 | 6 | 1 | 5 |
| Eval 2 | 4.3 | 4.7 | 13 | 6 | 7 | 5 | 1 | 4 |
| Eval 3 | 4.0 | 4.6 | 36 | 24 | 12 | 10 | 3 | 7 |
| Eval 4 | 3.1 | 4.3 | 81 | 40 | 41 | 32 | 10 | 22 |
| Average | 3.8 | 4.6 | 43 | 24 | 19 | 13 | 4 | 10 |

Table 6: Task 2 - Bilingual results for Overall Ratings and Error Annotation for the **Pairwise MT** scenario.

### 4.1 Are monolingual results comparable to bilingual results when evaluating a single MT?

**Overall Ratings -** Despite the variation in the overall Ratings, monolingual and bilingual evaluation of a single MT system seem to be comparable. The average scores of the four evaluators are close, at 3.7 for the monolingual task and 3.8 for the bilingual task, indicating that the monolingual evaluation may be as effective as the bilingual evaluation when evaluating a single MT.

**Error Annotation -** *All errors*: The results for the monolingual and bilingual tasks are relatively close when considering all errors tagged. The average number of all errors is 24 in the monolingual task, and 29 errors in the bilingual task, a difference of 17%. *Major errors*: Similarly, the results for the monolingual and bilingual tasks are relatively close when considering only Major errors tagged. The average results are close, with an average of 11 Major errors in the monolingual task, and 9 in the bilingual task.

These results show that both monolingual and bilingual evaluations are capturing enough severe errors to reflect the performance of one MT. This indicates that a monolingual evaluation may be effective when using the MQM typology, where *severity* is taken into account.

## 4.2 Are monolingual results comparable to bilingual results when evaluating two different MTs?

**Overall Ratings -** When evaluating two MT systems, both monolingual and bilingual evaluation seem to agree very closely. In both tasks, evaluators agree that MT2 performs well, with close average Ratings of 4.7, in the monolingual task, and 4.6 in the bilingual task. For MT1, even with a wider variation in Ratings, and also a stricter view of performance from evaluator 4, the average Ratings of four evaluators is very close, with 3.8 for both monolingual and bilingual. Both the monolingual and the bilingual evaluations established that MT2 outperforms MT1, indicating the usefulness of the monolingual evaluation.

**Error Annotation -** *All errors*: The results for the monolingual and bilingual tasks are close, with MT2 showing 15 errors in the monolingual and 13 in the bilingual tasks; and MT1 showing 43 errors in both monolingual and bilingual tasks. The difference of 2 errors in the MT2 assessment corresponds to 13% only. *Major errors*: Similarly, when only Major errors are considered, the results for the monolingual and bilingual tasks are close. The average number of Major errors tagged in MT2 is 5 for monolingual and 4 for bilingual, while in MT1 is 27 for monolingual and 24 for bilingual, also relatively close.

Both the monolingual and the bilingual evaluations are able to capture similar number of errors and severity, and establish that MT2 outperforms MT1. Both monolingual and bilingual are capturing enough severe errors to reflect the performance of the translation, indicating that a monolingual evaluation may be as effective as the bilingual evaluation in terms of Error Annotation for two MT systems.

## 4.3 Can monolingual assessments of MT be effective?

Following the results of the two evaluation tasks, we performed an analysis of the systems' outputs in order to identify what errors could and could not be detected monolingually, and also why they could be detected.

- **Most errors appear to be detectable monolingually (from target only)**

Analyzing all reported errors and categorizing them as either "detectable monolingually" or "not de-tectable monolingually" revealed that the vast majority (98%) seem to be identifiable solely from the target language (see Table 7). This finding aligns with previous results demonstrating that monolingual evaluation seems to yield comparable insights to bilingual evaluation.

| Monolingual | Detected | Not Detected | % Detected |
|---|---|---|---|
| Minor | 182 | 2 | 98.91% |
| Major | 143 | 6 | 95.97% |
| Total | 325 | 8 | 97.60% |

Table 7: Monolingual detectability results. "Detected" refers to errors identified monolingually, "Not Detected" refers to errors missed monolingually, and "% Detected" is the percentage of errors detected monolingually for each error type.

- **Many errors seem to be detectable from the context of the target only**

The impact of context in identifying errors is noticeable. One text from Test Set 2 (see Table 1) discusses the Great Wall of China. In Brazilian Portuguese (pt-BR), the word "wall" has three distinct translations: **parede** (for interior walls), **muro** (for perimeter walls around properties), and **muralha** (for defensive walls surrounding cities or fortresses, as in the case of the Great Wall). One of the MT systems inconsistently used all three translations, resulting in clear mistranslations that became particularly apparent when viewed in context:

*"Agora sobre o **muro** em si. Mutianyu é a seção mais longa e totalmente restaurada da Grande Muralha aberta aos turistas. Existem 23 torres de vigia, cerca de uma a cada cem metros em uma montanha ascendente e eu quero dizer realmente eles são íngremes no S\*\*\*! Perdoem a minha linguagem, mas maldita a maioria de nós estava cansada em cobrir apenas 5 desses. Ambos os lados da **muralha** têm um parapeito crenado para que os soldados pudessem disparar flechas contra o inimigo em ambos os lados. Isso é muito raro em outras seções da **parede**".*

Another type of error that became noticeable through context was grammatical gender. For example, in a text about the Notre Dame cathedral (Test Set 1), several mistranslations arose due to gender mismatches, as the word catedral ("cathedral") is feminine in Portuguese. In the example below, the pronoun ele (masculine) should have been ela (feminine) to correctly refer to Notre Dame:

*Para mim, a única atração absoluta de Paris era **Notre Dame** e eu nem tinha percebido. Fiquei impressionado com o detalhe e a sensação que **ele** deu!*

These errors were identifiable in the monolingual task due to the availability of context, showing that a significant number of errors seem to be detectable through monolingual evaluation when contextual information is present.

- **Some errors seem to be detectable only through bilingual evaluations**

Building on the previous findings (see Table 7), the analysis indicates that only a small proportion of errors require access to the source text for detection. While approximately 98% of errors were identified monolingually, a limited number of mistranslations remained undetected without bilingual evaluation. Examples include:

- The phrase "Find guest order details" was mistranslated as "Find guest order," omitting the key term "details".

- The source sentence "the mountain ridge was steep" was translated as "the mountain ridge was incredible", replacing "steep" with an incorrect adjective, which is not visible without the source.

- The sentence "They had some pan cake places too so it does have more than just Chinese menu" was mistranslated as "cake", entirely omitting the compound "pan cake."

- In the example, "You can usually buy on eBay without an account if the item is selling for less than $5,000 and it's offered with Buy It Now. You need an eBay account to bid on an auction item," the second sentence was omitted entirely from the translation.

In contrast, certain errors were evident from the target text alone, without requiring the source text for detection. For example:

- The sentence "how to buy as a guest on eBay" was translated as "Como comprar como hóspede no eBay." Here, the MT incorrectly translated "guest" as "hóspede" which refers to a hotel guest in Portuguese, instead of the correct term "convidado," meaning a guest for an event or website.

| Questions | Task 1 | Task 2 |
|---|---|---|
| Ease of finding error | 8.25 | 9.5 |
| Confidence in the Single MT eval | 8.0 | 9.25 |
| Confidence in the Pairwise MT eval | 8.0 | 9.0 |
| Time required | 3.75 | 6.5 |
| Effort required | 7.0 | 6.5 |
| Satisfaction with evaluation | 8.5 | 9.25 |

Table 8: Survey results for Monolingual (task 1) and Bilingual (task 2) evaluation.

- The phrase "Top Takeaway" was left untranslated, which is a clear error visible in the target language as it remains in English rather than the target language.

These findings show that while bilingual evaluation offers additional layers of verification, the majority of errors were identified through careful monolingual evaluation.

### 4.4 Survey

In order to grasp the evaluators' view on both tasks, monolingual and bilingual, a survey was designed. After each task, evaluators answered five questions, on a scale from 0-10 (where 0 is a low score, and 10 a high score) relating to the issues shown in table 8.[3] These results reflect the subjective perception of a limited number of evaluators and are intended to provide some basic information on the experience of the evaluators when doing evaluations in monolingual and bilingual scenarios.

The results suggest some differences between monolingual (Task 1) and bilingual (Task 2) evaluation approaches. Overall, evaluators found the bilingual evaluation easier, with higher scores for the ease of finding errors (9.5 vs 8.25) and greater satisfaction (9.25 vs 8.5). Confidence levels were slightly higher in the bilingual evaluation for single MT assessment (9.25 vs 8.0) and for comparing two MT outputs (9.0 vs 8.0).

Regarding the evaluators' note on time and effort, we note that bilingual evaluations may have required more time (6.5 vs 3.75) due to the need to read and compare the source text alongside the translations. Interestingly, despite taking more time, bilingual evaluations were perceived as requiring less effort (6.5 vs 7.0). This suggests that "less effort" in the bilingual modality may reflect reduced cognitive strain rather than overall time efficiency. Finally, while most results showed slight differences between tasks, a noticeable increase in time required for bilingual evaluations suggests a trade-off between effort and time efficiency. This finding suggests that less effort does not necessarily equate to less time spent on the task.

---
[3]See Appendix C for full questions.

## 5 Conclusion and Future work

This study, while limited in scope, highlights the potential of monolingual evaluation as a practical and effective alternative method for assessing MT performance. The results suggest that monolingual evaluation may be comparable to bilingual evaluation in assessing the performance of one MT output and in comparing multiple MT outputs - an approach commonly used in companies and academia alike.

Monolingual evaluation seems to be particularly effective when using contextual information, which reinforces the importance of assessing translations within context rather than isolated sentences. Monolingual evaluation also seems to show a robust capability for error detection, with approximately 98% of errors identifiable by examining the target language alone. These findings suggest the potential usefulness of monolingual approaches as both a complementary and standalone alternative method for MT evaluation. Additionally, findings from the survey suggest that monolingual evaluation may not significantly differ from bilingual evaluation in terms of translators confidence and satisfaction. Notably, monolingual evaluation may be a faster task. While evaluators reported slightly higher effort — likely due to the need for rereading in the absence of a source text — this increased cognitive effort does not translate to longer task completion times. As a result, monolingual evaluations may offer a more time- and cost-efficient approach to MT assessment without compromising reliability.

By suggesting the potential effectiveness of monolingual evaluation, this paper contributes to the field of MT evaluation and highlights potential benefits. One notable potential advantage is the expansion of the evaluator pool, as monolingual evaluations require proficiency in only one language, making it easier to find qualified evaluators, even for less commonly spoken language pairs. Furthermore, the reliance on monolingual assessment aligns well with the needs of generative AI evaluation, where systems produce text in a single language. This suggests that monolingual evaluation could serve as a valuable framework for assessing text generation tasks beyond MT, including AI-generated content.

## Acknowledgments

Our thanks to Katja Zuske for the review and to the evaluators that helped this effort. The second author benefits from being member of the ADAPT SFI Research Centre at Dublin City University, funded by the Science Foundation Ireland under Grant Agreement No. 13/RC/2106_P2.

## A  Annotation Guidelines

The Annotation Guidelines used are the ones published by (Freitag et al., 2021) and are displayed in table 9.

## B  Error Categories, Severity and Description

The Error Categories, along with their severity and description, were published by (Freitag et al., 2021) and are displayed in tables 10 and 11.

## C  Survey questions

The survey questions used can be seen in table 12.

You will be assessing translations at the segment level, where a segment may contain one or more sentences. Each segment is aligned with a corresponding source segment, and both segments are displayed within their respective documents. Annotate segments in natural order, as if you were reading the document. You may return to revise previous segments.

Please identify all errors within each translated segment, up to a maximum of five. If there are more than five errors, identify only the five most severe. If it is not possible to reliably identify distinct errors because the translation is too badly garbled or is unrelated to the source, then mark a single Non-translation error that spans the entire segment.

To identify an error, highlight the relevant span of text, and select a category/sub-category and severity level from the available options. (The span of text may be in the source segment if the error is a source error or an omission.) When identifying errors, please be as fine-grained as possible. For example, if a sentence contains two words that are each mistranslated, two separate mistranslation errors should be recorded. If a single stretch of text contains multiple errors, you only need to indicate the one that is most severe. If all have the same severity, choose the first matching category listed in the error typology (eg, Accuracy, then Fluency, then Terminology, etc).

Please pay particular attention to document context when annotating. If a translation might be questionable on its own but is fine in the context of the document, it should not be considered erroneous; conversely, if a translation might be acceptable in some context, but not within the current document, it should be marked as wrong.

There are two special error categories: Source error and Non-translation. Source errors should be annotated separately, highlighting the relevant span in the source segment. They do not count against the five-error limit for target errors, which should be handled in the usual way, whether or not they resulted from a source error. There can be at most one Non-translation error per segment, and it should span the entire segment. No other errors should be identified if Non-Translation is selected.

Table 9: MQM Annotator guidelines

| Error Category | Description |
| --- | --- |
| Accuracy | |
| -Addition | -Translation includes information not present in the source. |
| -Omission | -Translation is missing content from the source. |
| -Mistranslation | -Translation does not accurately represent the source. |
| -Untranslated text | -Source text has been left untranslated. |
| Fluency | |
| -Punctuation | -Incorrect punctuation (for locale or style). |
| -Spelling | -Incorrect spelling or capitalization. |
| -Grammar | -Problems with grammar, other than orthography. |
| -Register | -Wrong grammatical register (eg, inappropriately informal pronouns). |
| -Inconsistency | -Internal inconsistency (not related to terminology). |
| -Character encoding | -Characters are garbled due to incorrect encoding. |
| Terminology | |
| -Inappropriate for context | -Terminology is non-standard or does not fit context. |
| -Inconsistent use | -Terminology is used inconsistently. |
| Style | |
| -Awkward | -Translation has stylistic problems. |
| Locale convention | |
| -Address format | -Wrong format for addresses. |
| -Currency format | -Wrong format for currency. |
| -Date format | -Wrong format for dates. |
| -Name format | -Wrong format for names. |
| -Telephone format | -Wrong format for telephone numbers. |
| -Time format | -Wrong format for time expressions. |
| Other | -Any other issues. |
| Source error | -An error in the source. Non-translation Impossible to reliably characterize distinct errors. |
| Non-translation | -Impossible to reliably characterize distinct errors. |

Table 10: Error categories with their definitions

| Severity | Severity Definition |
|---|---|
| Major | Actual translation or grammatical errors |
| Minor | Smaller imperfections |
| Neutral | Purely subjective opinions about the translation |

Table 11: Severities and their definitions

| Questions |
|---|
| How easy was it to find errors in the monolingual evaluation? (0 not at all, 10 very easy) |
| How confident were you in your evaluation when assessing the SINGLE output monolingually? (0 not at all, 10 completely confident) |
| How confident were you in your evaluation when assessing the TWO MT outputs monolingually? (0 not at all, 10 completely confident) |
| How much time did the monolingual evaluation require? (0 very little time, 10 a lot of time) |
| How much effort did the monolingual evaluation require? (0 very little effort, 10 a lot of effort) |
| How satisfied were you with the monolingual evaluation? (0 not at all, 10 completely satisfied) |
| |
| How easy was it to find errors in the bilingual evaluation? (0 not at all, 10 very easy) |
| How confident were you in your evaluation when assessing the SINGLE output bilingually? (0 not at all, 10 completely confident) |
| How confident were you in your evaluation when assessing the TWO MT outputs bilingually? (0 not at all, 10 completely confident) |
| How much time did the bilingual evaluation require? (0 very little time, 10 a lot of time) |
| How much effort did the bilingual evaluation require? (0 very little effort, 10 a lot of effort) |
| How satisfied were you with the bilingual evaluation? (0 not at all, 10 completely satisfied) |

Table 12: Survey questions